\documentclass[twoside,leqno,twocolumn]{article}
\usepackage{ltexpprt}

\usepackage{amssymb,amsmath,dsfont}
\usepackage{bm}
\usepackage{algorithm, algorithmic}
\usepackage{graphicx}
\usepackage{subfigure,comment}
\usepackage{enumitem}
\usepackage{multirow}
\usepackage{chngpage}

\newcommand{\quartertab}{\hspace*{.5em}}
\newcommand{\eighthtab}{\hspace*{.25em}}

\DeclareMathOperator*{\argmax}{arg\,max}

\newcommand{\ie}{\textit{i.e.}}
\newcommand{\eg}{\textit{e.g.}}

\newtheorem{definition}{Definition}
\begin{document}

\title{\Large Detecting Unusual Input-Output Associations in Multivariate Conditional Data}
\author{Charmgil Hong\thanks{Department of Computer Science, University of Pittsburgh.} \\
\and
Milos Hauskrecht$^*$}
\date{}

\maketitle

%\pagenumbering{arabic}
%\setcounter{page}{1}%Leave this line commented out.

\begin{abstract} \small\baselineskip=9pt 
Despite tremendous progress in outlier detection research in recent years, the majority of existing methods are designed only to detect \textit{unconditional} outliers that correspond to unusual data patterns expressed in the joint space of all data attributes.
Such methods are not applicable when we seek to detect \textit{conditional} outliers that reflect unusual responses associated with a given context or condition.
This work focuses on \textit{multivariate conditional outlier detection}, a special type of the conditional outlier detection problem, where data instances consist of multi-dimensional input (context) and output (responses) pairs.
We present a novel outlier detection framework that identifies abnormal input-output associations in data with the help of a decomposable conditional probabilistic model that is learned from all data instances.
Since components of this model can vary in their quality, we combine them with the help of weights reflecting their reliability in assessment of outliers.
We study two ways of calculating the component weights: global that relies on all data, and local that relies only on instances similar to the target instance.
Experimental results on data from various domains demonstrate the ability of our framework to successfully identify multivariate conditional outliers.
\end{abstract}

% INTRODUCTION
\section{Introduction}

\textit{Outlier detection} is a data analysis task that aims to find atypical behaviors, unusual outcomes, erroneous readings or annotations in data.\footnote{\textit{Outliers} are also referred to as \textit{anomalies}, \textit{abnormalities}, \textit{novelties}, \textit{discordances}, or \textit{deviants}.}
It has been an active research topic in data mining community, and it is frequently used in various applications to identify rare and interesting data patterns, which may be associated with beneficial or malicious events, such as
fraud identification \cite{Wang:2010:ICICTA}, %Fawcett:1997,Bolton:2002
network intrusion surveillance \cite{garcia:2009}, %Tan:2002,Zhang:2010:IET
disease outbreak detection \cite{Wong:2003:ICML},
patient monitoring for preventable adverse events (PAE) \cite{Hauskrecht:2007,Hauskrecht:2013},
%astronomical data analysis \cite{zhang:2004:SPIE,borne:2009:bookchapter}, %ball:2010:IJMP
%financial market monitoring and analysis \cite{tolvi:2003,cao:2005}, %chong:2005
\textit{etc}.
It is also utilized as a primary data preprocessing step that helps to remove noisy or irrelevant signals in data \cite{Hodge:2004:SOD,Liu:2004:CCE}. %Witten:2011:DMP,aggarwal:2013:book,Pan:2016:CVPR

% global climate monitoring and sudden change/natural disaster detection \cite{peterson:1998:IJC,Lakshmanan:2000,sun:2012,rahmani:2014},
% genetic diversity analysis and new species discovery in taxonomic studies \cite{foll:2008,chen:2009:PAMI,coop:2010,cullingham:2014}, abnormal energy consumption detection \cite{li:2010:IE,seem:2007,branch:2013:kis}, Automatic Identification System (AIS) tracking in vessel traffic services \cite{mascaro:2010:tr,pallotta:2013,handayani:2013}, %kamstra:2000

Despite an extensive research, the majority of existing outlier methods are developed to detect \textit{unconditional} outliers that are expressed in the joint space of all data attributes.
Such methods may not work well when one wants to identify \textit{conditional} (contextual) outliers that reflect unusual responses for a given set of contextual attributes. Briefly, since conditional outliers depend on the context or properties of data instances, application of unconditional outlier detection methods may lead to incorrect results.
For example, assume we want to identify incorrect (or highly unusual) image annotations in a collection of annotated images. 
Then by applying unconditional detection methods to the joint image-annotation space may lead to images with rare themes to be falsely identified as outliers due to the scarcity of these themes in the dataset, leading to false positives. 
Similarly, an unusual annotation of images with frequent themes may not be judged (scored) as very different from images with less frequent themes leading to false negatives.  
%%For example, consider a collection of annotated images where one attempts to find incorrect image annotations using an unconditional detection method.
%%Images with rare subjects may be falsely identified as outliers due to the scarcity of the subjects in the dataset ({false positives}). 
%%%% Similarly, in hospitals, a moderately high medication dose, which is common when considered for the overall patient population, is %%% unusual and concerning if it is administered to a pediatric patient ({false negative}).
%%%% Choose one examples
%%%% Similarly, while an image with a giraffe label appears normal with respect to all images, an image of an ocean with an object that is annotated as a giraffe is unusual given the ocean context and hence could be missed by unconditional methods ({false negatives}).  

This paper focuses on \textit{multivariate conditional outlier detection}, a special type of the conditional outlier detection problem where data consists of $m$-dimensional continuous input vectors (context) and corresponding $d$-dimensional binary output vectors (responses).
Our goal is to precisely identify the instances with unusual input-output associations.
Following the definition of outlier given by Hawkins \cite{hawkins:1980:book}, we give a description of multivariate conditional outlier in plain language as:
\begin{definition}
A multivariate conditional outlier is an observation, which consists of context and associated responses, whose responses are deviating so much from the others in similar contexts as to arouse suspicions that it was generated by a different response mechanism.
\end{definition}
This formulation fits well various practical outlier detection problems that require contextual understanding of data.
As briefly illustrated above, for example, recent social media services allow users to tag their content (\eg, online documents, photos, or videos) with keywords and thereby permit keyword-based retrieval.
These user annotations sometimes include irrelevant words by mistake that could be effectively pinpointed if the conditional relations between content and tags are considered.
Likewise, evidence-based expert decisions (\eg, functional categorization of genes, medical diagnosis and treatment decisions of patients) occasionally involve errors that could cause critical failures.
Such erroneous decisions would be adequately detected through contextual analysis of evidence-decision pairs.

The multivariate conditional outlier detection problem is challenging because both the contextual- and inter-dependences of data instances should be taken into account when identifying outliers. 
We tackle these challenges by building a probabilistic model $P(\mathbf{Y}|\mathbf{X})$, where $\mathbf{X}\!=\!(X_1, ..., X_m)$ denotes the input variables and $\mathbf{Y}\!=\!(Y_1, ..., Y_d)$ denotes the associated output variables. 
Briefly, the model is built (learned) from all available data, aiming to capture and summarize all relevant dependences among data attributes and their strength as observed in the data. 
Conditional outliers are then identified with the help of this model. 
More specifically, a conditional outlier corresponds to a data instance that is assigned a low probability by the model.

The exact implementation of the above approach is complicated, and multiple issues need to be resolved before it can be applied in practice.
First, it is unclear how the probabilistic model $P(\mathbf{Y}|\mathbf{X})$ should be represented and parameterized.
To address this problem, we resort to and adapt structured probabilistic data models of $P(\mathbf{Y}|\mathbf{X})$ that provide an efficient representation of input-output relations by decomposing the model using the chain rule into a product of univariate probabilistic factors $P(Y_i|\mathbf{X},\mathbf{Y}_{\boldsymbol{\pi}(i)}):i\!=\!1,...,d$; 
\ie, each response $Y_i$ is dependent on $\mathbf{X}$ and a subset of the other responses $\mathbf{Y}_{\boldsymbol{\pi}(i)}$. 
The univariate conditional models and their learning are rather common and well studied, and multiple models (\eg, logistic regression) can be applied to implement them. 
We note the structured probabilistic data models were originally proposed and successfully applied to support structured output prediction problems \cite{Zhang:2013}. 
However, their application to outlier detection problems is new. 
The key difference is that while in prediction we seek to find outputs that maximize the probability given the inputs, in conditional outlier detection we aim to identify unusual (or low probability) associations in between observed inputs and outputs. 

The second issue is that the probabilistic model must be learned from available data which can be hard especially when the number of context and output variables is high and the sample size is small. 
This may lead to model inaccuracies and miscalibration of probability estimates, which in turn may effect the identification of outliers. 
To alleviate this problem, we formulate and present outlier scoring methods that combine the probability estimates with the help of weights reflecting their reliability in assessment of outliers.

%Through empirical studies on datasets with multi-dimensional responses, we demonstrate that our method is able to successfully identify multivariate conditional outliers.
Through empirical studies, we test our approach on datasets with multi-dimensional responses. 
We demonstrate that our method is able to successfully identify multivariate conditional outliers and outperforms the existing baselines. 

The rest of this paper is organized as follows.
Section \ref{sec:problem_def} formally define the problem. 
Section \ref{sec:related} reviews existing research on the topic.
Section \ref{sec:approach} describes our multivariate conditional outlier detection approach.
Section \ref{sec:experiments} presents the experimental results and evaluations.
Lastly, Section \ref{sec:concl} summarizes the conclusions of our study.% and discusses possible future research directions.

% PROBLEM DEFINITION
\section{Problem Definition}
\label{sec:problem_def} 

In this work, we study a special type of the conditional outlier detection problem where data consist of multi-dimensional input-output pairs; 
that is, each instance in dataset $\mathcal{D}\!=\!\{\mathbf{x}^{(n)}, \mathbf{y}^{(n)}\}_{n=1}^N$ consists of an $m$-dimensional continuous input vector $\mathbf{x}^{(n)}\!=\!(x_1^{(n)}, ..., x_m^{(n)})$ and a $d$-dimensional binary output vector $\mathbf{y}^{(n)}\!=\!(y_1^{(n)}, ..., y_d^{(n)})$.
%Our goal is to detect instances with unusual input-output associations in $\mathcal{D}$.
%We assume that there are deterministic response processes from input to output (\eg, image tagging or medication prescription as in the previous examples) and aim to detect irregular response patterns in $\mathbf{Y}$ given context $\mathbf{X}$.
%%% The outputs are linked to inputs via different processes (\eg, image tagging or medication prescription as in the previous examples).  
Our goal is to detect irregular response patterns in $\mathbf{Y}$ given context $\mathbf{X}$.
The fundamental issues in developing a multivariate conditional outlier detection method are how to take into account the \textit{contextual dependences between output $\mathbf{Y}$ and their input $\mathbf{X}$}, as well as the \textit{mutual dependences among $\mathbf{Y}$}. 
We address these issues by building a decomposable probabilistic representation for $\mathbf{Y|X}$. 

Note that multivariate conditional outlier detection is clearly different from unconditional outlier detection when the problems are expressed probabilistically.
In conditional outlier detection, we are interested in the instances that fall into low-probability regions of the conditional joint distribution $P(\mathbf{y}|\mathbf{x})\!=\!P(\mathbf{y},\mathbf{x})/P(\mathbf{x})$.
On the other hand, unconditional outlier detection approaches generally seek instances in low-probability regions of the joint distribution $P(\mathbf{y},\mathbf{x})$.

\vspace{.5em}
\begin{scriptsize}
\noindent
{\textbf{Notation:} For notational convenience, we will omit the index superscript $^{(n)}$ when it is not necessary. We may also abbreviate the expressions by omitting variable names; e.g., $P(Y_1\!=\!y_1, ..., Y_d\!=\!y_d|\mathbf{X\!=\!x}) = P(y_1, ..., y_d|\mathbf{x})$.}
\end{scriptsize}

% RELATED RESEARCH + PROB DEF
\section{Existing Research}
\label{sec:related}

Outlier detection has been extensively studied in the data mining and statistics communities \cite{Chandola:2009:ACMCS,kriegel:2010:sdm,aggarwal:2013:book}. 
A wide variety of approaches to tackle the detection problem for multivariate data have been proposed in the literature.    
%Despite its long research history, however, the concept of outlier is rather ill-defined and, indeed, there is no clear consensus on what an outlier is.
%Instead, each approach and method has been proposed with its own definition and assumptions of outliers.
Accordingly, depending on the type of outliers the method aims to detects, five general categories of \textit{unconditional} outlier detection approaches appear in the literature.
These include 
density-based approaches \cite{breunig:2000,papadimitriou:2003:ICDE}, %jin2001mining,Zhang:1996:BED:235968.233324
distance-based approaches \cite{Rousseeuw:1990,Bay:2003:KDD}, %,Knorr:1997:CASCON,Angiulli:2002:FOD:645806.670167,Knorr97aunified,Fan_anonparametric}, 
depth-based approaches \cite{Rousseeuw:1987:RRO,scholkopf:1999:NIPS}, %Tukey:1977:book,Johnson98fastcomputation,,amer:2013:kdd
deviation-based approaches \cite{Arning:1996:KDD}, 
and high-dimensional approaches \cite{Aggarwal:2001:SIGMOD,kriegel:2010:sdm}.  %zou:2006:sparse
%Globerson:2003:UAI,Mohamed:2008:NIPS,huang:2012:ejs,lawrence:2009, Ruts:1996:CDC:255810.255825,
Below we briefly summarize each of these categories.
For technical details, please refer to \cite{Chandola:2009:ACMCS,kriegel:2010:sdm,aggarwal:2013:book}.

%One of the most referred definition, though, would be: 
%\textit{``an outlier is an observation deviating so much from the others as to arouse suspicions that it was generated by a different mechanism''} \cite{hawkins:1980:book}.
%Within this rather broad definition, a variety of ideas and techniques have been proposed to identify irregular deviations in multivariate data space \cite{kriegel:2010:sdm}.

Density-based approaches assume that the density around a normal data instance is similar to that of its neighbors \cite{breunig:2000,papadimitriou:2003:ICDE}.
A typical representative method is Local Outlier Factor (LOF) \cite{breunig:2000}, which measures a relative local density in $k$-nearest neighbor boundary.
%An LOF score indicates the relative rarity of an instance ...
%influenced several subsequent works in the literature \cite{papadimitriou:2003:ICDE}.
LOF has shown good performance in many applications and is considered as an off-the-shelf outlier detection method.
In Section \ref{sec:experiments}, we use LOF as the representative unconditional outlier detection method and compare the performance with our proposed approach.

Distance-based approaches assume that normal data instances come from dense neighborhoods, while outliers correspond to isolated points.
A representative method is \cite{Rousseeuw:1990} which gives an outlier score to each instance using a robust variant of the Mahalanobis distance \cite{Rousseeuw:1984}, measuring the distance between each instance to the main body of data distribution such that the instances located far from the center of data distribution are identified as outliers.
%Other methods that fall in this category include Knorr's unified approach \cite{Knorr:1997:CASCON}, randomized pruning method \cite{Bay:2003:KDD}, \textit{etc}.
%linearization method \cite{Angiulli:2002:PKDD},  resolution based method \cite{Fan:2006:PAKDD}, 
%One limitation of the distance-based approaches is that they suffer from the ``curse of dimensionality'' issue which makes them less suitable for high-dimensional data.

Depth-based approaches assume that outliers are at the fringe of the data regions and normal instances are close to or in the center of the region.
The methods in this category assign depth $k$ to each instance by gradually removing data from convex hulls, and the instances with small depth are considered as outliers \cite{Rousseeuw:1987:RRO}. 
A relevant method is the One-class Support Vector Machines \cite{scholkopf:1999:NIPS}, which assumes all the training data belong to the ``normal'' class and finds a decision boundary defining the region of normal data, whereas instances lie across the boundary are identified as outliers.

Deviation-based approaches assume that outliers are the outmost data instances in the data region and can be identified by measuring the impact of each instance on the variance of the dataset.
%Accordingly, it measures how much the variance in the dataset is reduced (so-called smoothing factor) when a particular set of data instances is removed.
One of the well-known algorithms in this category is Linear Method for Deviation Detection (LMDD) \cite{Arning:1996:KDD}.
Compared to the depth-based approaches, deviation-based approaches do not require complicated contour generation process.
% need to assign the depth $k$, nor
%Also, since the approach is distribution-independent, it is widely applicable.
%However, for sample size $N$, there are $2^N$ options of which instances to remove, which makes the approaches less applicable to large datasets.

%kriegel:2010:sdm
In high-dimensional spaces, the above approaches often fail because the distance metrics and density estimators become computationally intractable and analytically ineffective.
Moreover, due to the sparsity of data, no meaningful neighborhood can be defined.
High-dimensional approaches are proposed to handle such extreme cases.
Typical methods in this category project the data to a lower dimensional subspace, 
such as grid-based subspace outlier detection \cite{Aggarwal:2001:SIGMOD}.
For a detailed review on related methods, see \cite{kriegel:2010:sdm}. 
%sufficient dimensionality reduction \cite{Globerson:2003:UAI}, Bayes Exponential Family PCA \cite{Mohamed:2008:NIPS}, 
%and Sparse PCA \cite{zou:2006:sparse}. More recent methods use Gaussian processes to help matrix factorization \cite{lawrence:2009}, 
%explore the structure between the data instances \cite{huang:2012:ejs},
%or define non-parametric outlier score functions \cite{hero:2006:nips,zhao:2009:nips}. %scott:2006:jmlr.

%does not have a good natural fit to any of 
While the vast majority of existing work were built to solve the unconditional outlier detection problem, the approaches may not work properly when it comes to \textit{conditional} outliers, since they do not take into account the conditional relations among data attributes.
Realizing this, recent years have seen increased interest in the \textit{conditional} outliers detection that aims to identify outliers in a set of outputs for given values of inputs.
Several approaches have been proposed to address the problems in this regard \cite{Hauskrecht:2007,Hauskrecht:2013,Song:2007:TKDE}.
However, these solutions either are limited to handle problems with a single output variable \cite{Hauskrecht:2007,Hauskrecht:2013} or 
assume a restricted relations among real-valued input and output variables through a Gaussian mixture \cite{Song:2007:TKDE}.
As results, the existing methods either make an independence assumptions that is too restrictive or are unfit for modeling multi-dimensional binary output variables.

%Even if they are applied to our target problem, 
%As a result, although the methods may effectively detect outliers in a single response dimension, the outliers defined jointly across multiple dimensions may easily be missed. 
%Also, the methods may be robust to the outliers occurring in multiple response dimensions but less sensitive to the sparse outliers that are observed in one or only few response dimensions.
%Consequently, the solutions are not able to fully exploit the properties of the multivariate conditional problem.

In contrast to the existing methods, our proposed approach is different in that
(1) it properly models multi-label binary outputs by adopting a structured probabilistic data model to represent data; and
(2) it utilizes the decomposed conditional probability estimates from individual response dimensions to identify outliers.
Consequently, our proposed approach drives the process of outlier detection to a more granular level of the conditional behaviors in data and (as follows in Section \ref{sec:experiments}) leads to a significant performance improvement in outlier detection.
Furthermore, by maintaining separate models for individual output variables, our approach provides a practical advantage that the existing multivariate outlier detection methods do not allow. 
That is, one can delve into a trained multivariate conditional model and investigate the quality of each univariate representation $P(y_i|\mathbf{x},\mathbf{y}_{\boldsymbol{\pi}(i)})$ to decide whether the individual model could be reliably used to support outlier detection. For example, a univariate model that produces inconsistent estimates could be preemptively excluded from the outlier detection phase. Since our goal is not to recover a complete data representation but to obtain a useful utility function for outlier detection, this sort of modularity allows us to utilize only the model with high confidence and, hence, to perform more robust outlier detection.

% OUR APPROACH
\section{Our Approach}
\label{sec:approach}

This section describes our approach to identify unusual input-output pairs, which we refer to as MCODE: \textit{Multivariate Conditional Outlier DEtection}.
To facilitate an effective detection method, we utilize a decomposable probabilistic data representation for $P(\mathbf{Y}|\mathbf{X})$ to capture the dependence relations among inputs and outputs, and to assess outliers by seeking low-probability associations between them.
Accordingly, having a precise probabilistic data model and proper outlier scoring methods is of primary concern.
In Section \ref{subsec:approach_model}, we discuss how to obtain an efficient data representation and accurate conditional probability estimates of observed input-output pairs, using the probabilistic structured data modeling approach \cite{Read:2009:ECML}
In Section \ref{subsec:approach_score}, we treat the probability estimates as a proxy representation of observed instances and present two outlier scoring methods by analyzing the reliability of these estimates.

\subsection{Probabilistic Modeling and Estimation}
\label{subsec:approach_model}

Our MCODE approach works by analyzing data instances come in input-output pairs with a statistical model representing the conditional joint distribution $P(\mathbf{Y}|\mathbf{X})$. 
A direct learning of the conditional joint from data, however, is generally very expensive or even infeasible, because the number of possible output combinations grows exponentially with $d$.
To avoid such a high cost of learning yet achieve an accurate data representation for outlier detection, we decompose the conditional joint into a product of conditional univariate distributions using the chain rule of probability:
\begin{align}
\label{eq:chain}
P(Y_1,...,Y_d | \mathbf{X}) 
%	&= \prod_{i=1}^d P(Y_i|\mathbf{X},Y_1, ..., Y_d)\\
	&= \prod_{i=1}^d P(Y_i|\mathbf{X},\mathbf{Y}_{\boldsymbol{\pi}(i)})
\end{align}
where $\mathbf{Y}_{\boldsymbol{\pi}(i)}$ denotes the parents of $Y_i$; \ie, all the output variables preceding $Y_i$ \cite{Read:2009:ECML}.
This decomposition lets us represent $P(\mathbf{Y}|\mathbf{X})$ by simply specifying each univariate conditional factor, $P(Y_i|\mathbf{X},\mathbf{Y}_{\boldsymbol{\pi}(i)})$.
In this work, we use a logistic regression model for each of the output dimensions, because it can effectively handle high-dimensional feature space defined by a mixture of continuous and discrete variables (\ie, $\mathbf{X},\mathbf{Y}_{\boldsymbol{\pi}(i)}$ conditioning $Y_i$) using regularization \cite{ng:2004:icml,cetin:2001:ip}.\footnote{Depending on data types and assumptions, different probabilistic classification functions could also be used, \eg, na\"{i}ve Bayes, relevance vector machine, or probabilistic support vector machine.}

In theory, the result of the above product should be invariant regardless of the chain order (order of $Y_i$).
Nevertheless, in practice, different chain orders produce different conditional joint distributions as they draw in models learned from different data \cite{Dembczynski:2010:ICML}. 
For this reason, several structure learning methods that determine the optimal set of parents have been proposed \cite{Zhang:2010:KDD,Kumar:2012:ECML}. %Zaragoza:2011:IJCAI,
However, these methods require at least $O(d^2f_c)$ of time, where $f_c$ denotes the time of learning a classifier, that would not be preferable, especially when the output dimensionality $d$ is high.

In MCODE we address the above problem by relaxing the chain rule and by permitting circular dependences among the output variables.
That is, we let $\mathbf{Y}_{\boldsymbol{\pi}(i)}$, the parents of $Y_i$, be all the remaining output variables, and assume the true dependence relations among them could be recovered through a proper regularization of logistic regression. 
To summarize, our structural decomposition allows us to capture the interactions among the output variables, as well as the input-output relations, using a collection of individually trained probabilistic functions with a relaxed conditional independence assumption.
We use $\mathcal{M} = \{\theta_{\mathcal{M}(1)}, ..., \theta_{\mathcal{M}(d)}\}$ to denote this structured data representation, where $\theta_{\mathcal{M}(i)}$ is the parameters of the probabilistic model for the $i$-th output dimension. Assuming logistic regression, these base statistical functions are parameterized using $\mathcal{D}$ as:
\begin{align}
	\theta_{\mathcal{M}(i)} = \argmax_\theta \sum_{n=1}^N \log P(y_i^{(n)}|\mathbf{x}^{(n)},\mathbf{y}_{-i}^{(n)}; \theta)
\end{align}
This defines a pseudo-conditional joint probability of an observation pair $(\mathbf{x},\mathbf{y})$ as:
\begin{align}
\label{eq:chain_relax}
%\Psi(Y_1,...,Y_d | \mathbf{X};\mathcal{M}) &= \prod_{i=1}^d P(Y_i|\mathbf{X},\mathbf{Y}_{-i}; \theta_{\mathcal{M}(i)})
\Psi(y_1,...,y_d | \mathbf{x};\mathcal{M})
	&= \prod_{i=1}^d \widetilde{P}(y_i|\mathbf{x},\mathbf{y}_{-i}; \theta_{\mathcal{M}(i)})
\end{align}
where $\mathbf{y}_{-i}$ denotes the values of all other output variables except $Y_i$.

\begin{figure}[t]
\centering
	\includegraphics[width=0.45\textwidth]{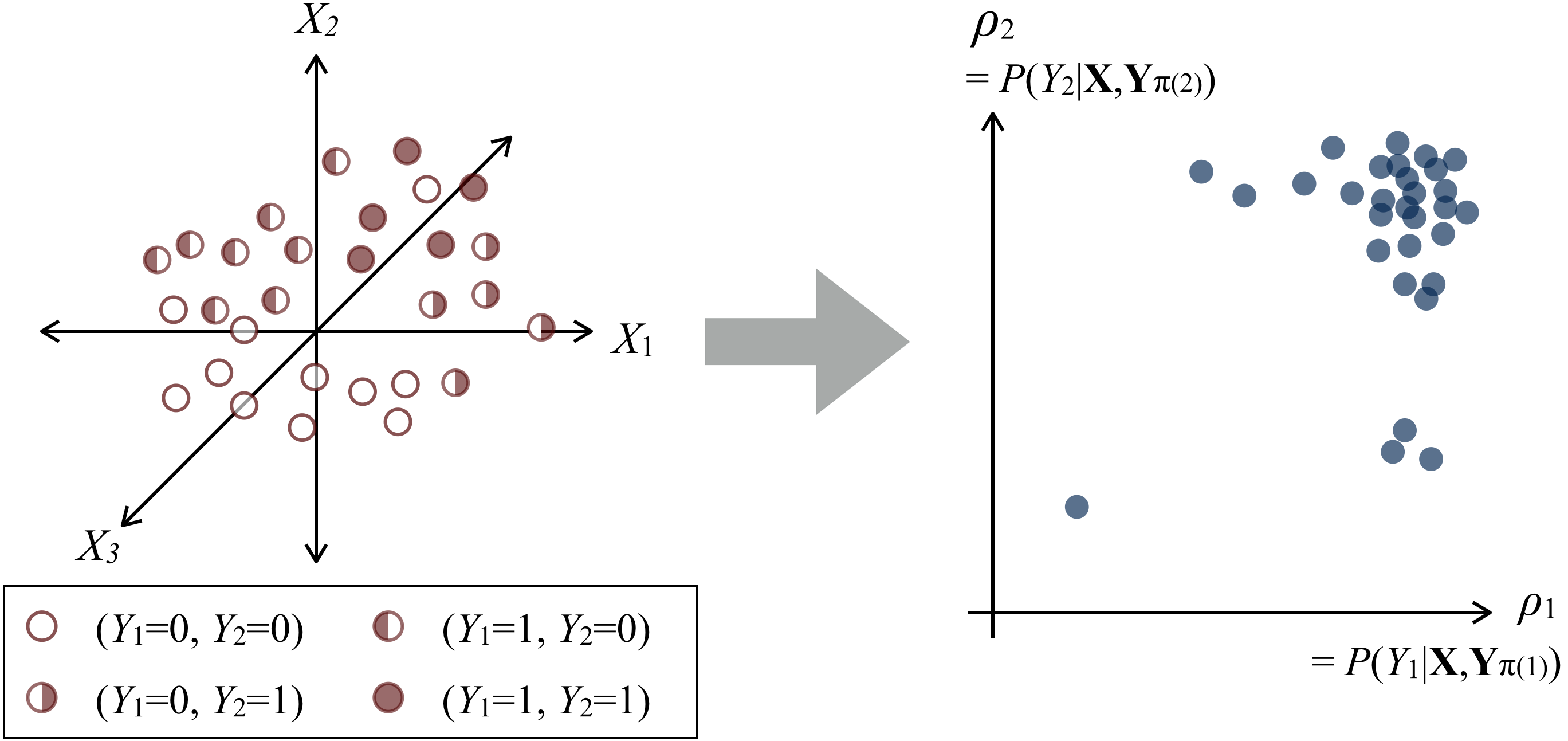}
	\caption{An illustrative example of conditional probability estimation process.}%Transformation of data from the original space (left) to the $d$-dimensional conditional likelihood space $\boldsymbol{\mathcal{P}}$ (right).}
	\label{fig:concept}
\end{figure}

Now let us apply our data representation $\mathcal{M}$ to estimate the conditional probabilities of observed outputs. 
For notational convenience, we introduce an auxiliary vector $\boldsymbol{\rho}\!=\!(\rho_1, ..., \rho_d)$ of $d$ random variables, each defined in a conditional probability space $\rho_i = [0,1]$.
%$(\Omega_i, \mathcal{F}_i, \mathcal{P}_i):i=1,...,d$. 
%Here, $\Omega_i = \{0,1\}$ is the sample space, equipped with a $\sigma$-algebra $\mathcal{F}_i$ and probability measure $\mathcal{P}_i:\Omega_i\rightarrow[0,1]$.
Each element of $\boldsymbol{\rho}$ is quantized by a probabilistic estimation process that is formalized as below by unleashing the product in Equation (\ref{eq:chain_relax}): 
\begin{align}
\mathcal{M}: \eighthtab(\mathbf{x}^{(n)},\mathbf{y}^{(n)}) \quartertab\rightarrow\quartertab\boldsymbol{\rho}^{(n)} 
				& = (\rho_1^{(n)}, ..., \rho_d^{(n)} ) %\in \boldsymbol{\mathcal{P}_\mathcal{M}}
\end{align}
where %$\rho_i^{(n)} = \widetilde{P}(y_i^{(n)}|\mathbf{x}^{(n)},\mathbf{y}_{-i}^{(n)}; \theta_{\mathcal{M}(i)})$.
$$
\rho_i^{(n)} = 
	\begin{cases}
		\eighthtab \widetilde{P}(y_i^{(n)}|\mathbf{x}^{(n)},\mathbf{y}_{-i}^{(n)}; \theta_{\mathcal{M}(i)}) &\text{ if }\quartertab y_i^{(n)} = 1\\[.75em]
		\eighthtab 1 - \widetilde{P}(y_i^{(n)}|\mathbf{x}^{(n)},\mathbf{y}_{-i}^{(n)}; \theta_{\mathcal{M}(i)}) &\text{ otherwise.}
	\end{cases}
$$
Accordingly, space of $\boldsymbol{\rho}$ is projecting a normalized confidence level (\ie, conditional probability estimate) of each observation $(\mathbf{x}^{(n)},\mathbf{y}^{(n)})$ across individual output dimensions, using the data representation $\mathcal{M}$.
Figure \ref{fig:concept} shows an illustrative example of this estimation where the input-output data instances (left) are projected to a 2-dimensional conditional probability space (right).

\subsection{Outlier Scoring}
\label{subsec:approach_score}

After the above probabilistic estimation process using $\mathcal{M}$, we consider the resultant conditional probabilities $\boldsymbol{\rho}$ as proxies of the original instances, and further hypothesize that multivariate conditional outliers could be effectively detected in this proxy space where instances are analyzed and expressed in terms of univariate posterior probabilities.
%In this respect, our approach can be understood as a reduction of multivariate conditional outlier detection to an (multivariate) unconditional outlier detection problem; 
Our goal is now to define an \textit{outlier score} that measures how unusual each input-output association is.

The most straightforward approach to define an outlier score is to use the probability $P(\mathbf{y}|\mathbf{x})$ of data instances calculated by the model $\mathcal{M}$:
\begin{align}
\label{eq:oscore_prod}
Score_\text{PROD}(\boldsymbol{\rho}^{(n)}) &= -\sum_{i=1}^d \log \rho_i^{(n)}
\end{align}
Please note that this assumes all probability estimates and the models generating them are of high quality.
However, in practice, the models that produce the probability estimates (\ie, $\theta_{\mathcal{M}(i)}$ in Equation (\ref{eq:chain_relax})) may not be all equally reliable as they are trained from a finite number of samples (this is important especially when the number of input and output variables is high, and the sample size is small).
Also, some dimensions of $Y_i|\mathbf{X},\mathbf{Y}_{\pi(i)}$ may not fit well the base statistical assumption (which in this work is a logistic curve) and result in miscalibrated estimations.
%Figure \ref{} shows such a discrepancy between dimensions. It plots the histograms of two dimensions of $\rho$
Consequently, if we treat all dimensions of $\boldsymbol{\rho}$ equally and merely search for the regions with low probabilities, the resulting scores degenerate to a noisy vector, which makes the detection of true irregularities hard.

To alleviate the issues, we propose to consider the reliability of each estimate dimension in $\boldsymbol{\rho}$ (\ie, the quality of model $\theta_{\mathcal{M}(i)}$) and adjust their influence in outlier scoring by weights that reflect their reliability.
We formalize our outlier score as:
\begin{align}
	\label{eq:oscore}
	Score_\text{RW}(\boldsymbol{\rho}^{(n)}) &= -\sum_{i=1}^d w_i \log \rho_i^{(n)}
\end{align}
where $w_i$ denotes the reliability weight of the model built for the $i$-th dimension.
Note that, when $w_i = 1$ for all dimensions $i\!=\!1,...,d$, the score becomes equivalent to Equation (\ref{eq:oscore_prod}), the negative log of the pseudo-conditional joint probability.

\subsubsection{Reliability Weights}

One way to define reliability weights would be to use the Brier score \cite{brier:1950} that measures the quality of the model in terms of model's probability outputs.
The Brier score is defined by averaging the squared errors of the probability estimates over all data instances:
\begin{align*}
	\frac{1}{N}\sum_{n=1}^N (f^{(n)} - o^{(n)})^2
%	\frac{1}{N}\sum_{N} \left(\text{(predicted probability)} - \text{(actual outcome)}\right)^2
%	\\	&= N^{-1}\sum_{n=1}^N (1-\rho_i^{(n)})^2
\end{align*}
where $f^{(n)}$ and $o^{(n)}$ respectively denotes the predicted probability and actual outcome of the $n$-th instance.
However, the assessment of the model quality for weighting purposes (Equation (\ref{eq:oscore})) by the Brier score may not be the best as the score imposes different penalties for different errors (the mean squared error penalizes larger errors more than smaller errors) 
and varies the distribution of errors \cite{willmott:2005:cr}. To address this, we propose our reliability weight be based on the mean estimated error, which gives the equal penalty to all errors:
%To address this, we propose a more refined reliability weight that measures and quantifies the distribution of errors in terms of its quantiles.
%To measure the dimension-wise reliability and give a proper set of weights to the scoring function, we utilize the quantile function of estimation errors.
\begin{definition}
Without loss of generality, let $\epsilon_i^{(n)} = 1-\rho_i^{(n)}$ be the estimated error of probability of an instance on dimension $i$. 
%Also, let $r_i$ denote the estimated reliability for the dimension.
Reliability weight $w_i$ is defined by taking the inverse of the mean estimated error:
\begin{align}
\label{eq:weight}
w_i = \frac{N}{\sum_{n=1}^N \epsilon_i^{(n)}}
\end{align}
\end{definition}
Our variant of the Brier-like score estimates the quality of each estimate dimension $\rho_i$ without distorting the distribution of errors. 
By taking the inverse of the score, we can effectively assign reliability weights to the dimensions, such that more on reliable dimensions become more important and the influence of noisy (unreliable) dimensions for outlier scoring is reduced.

% OSCORE 2
\subsubsection{Local Reliability Weights}

Notice that the above weighting scheme (Equation (\ref{eq:weight})) implicitly assumes that the reliability of probability estimates (\ie, the quality of a model) is invariant across all data regions.
However, the assumption often does not hold because in most practical problems especially with high-dimensional data spaces, data is not uniformly distributed in its attribute space.
That is, modeling and estimation of $P(Y_i|\mathbf{X},\mathbf{Y}_{\pi(i)})$ cannot be achieved properly in sparse regions of the attribute space. 

We tackle such a sparsity issue by evaluating the reliability of each dimension of $\boldsymbol{\rho}$ locally in the region around the instance we want to check.
This localized approach can be implemented as follows:
\begin{align}
	\label{eq:oscore_local}
	Score_\text{LRW}(\boldsymbol{\rho}^{(n)}) &= -\sum_{i=1}^d w_i^{(n)} \log \rho_i^{(n)}
\end{align}
where
\begin{align}
w_i^{(n)} = \frac{|N_k(n)|}{\sum_{n \in N_k(n)} \epsilon_i^{(n)}}
\end{align}
and $N_k(n)$ denotes $k$-nearest neighbors of the $n$-th instance in the original attribute space.
In the next section, we show the benefits of our reliability weights and outlier scores through experimental results.

% EXPERIMENTS
\section{Experiments}
\label{sec:experiments}

To validate and demonstrate the performance of our MCODE approach, we conduct experiments with data obtained from various domains.
Through the empirical analysis in this section, we would like to verify the advantages of 
(1) adopting the conditional outlier detection approach,
(2) considering the dependence relations among outputs,
(3) weighting via reliability estimation, and
(4) local reliability estimates and local outlier scores.
Below we describe our experimental design and present the evaluation results.

\subsection{Compared Methods}
\label{subsec:exp_methods}

To achieve our objectives, we perform experiments with the following methods:
\begin{itemize}[leftmargin=*]
	\setlength\itemsep{-0.1em}
	% LOF
	\item{\textit{Local outlier factor} (LOF) \cite{breunig:2000} -- 
	LOF is an unconditional method that estimates outliers using a relative local density measure in the joint space of all data attributes:
	\begin{align*}
		LOF\left((\mathbf{x,y}),k\right) &=  \frac{\sum_{(\mathbf{x',y'}) \in N_k(\mathbf{x,y})}\frac{\textit{lrd}_k (\mathbf{x',y'})}{\textit{lrd}_k (\mathbf{x,y})}}{\left| N_k(\mathbf{x,y}) \right|}
	\end{align*}
	where $N_k(\mathbf{x,y})$ denotes the $k$-nearest neighborhood of instance $(\mathbf{x,y})$ and
	\begin{align*}
		\textit{lrd}_k (\xi) = \frac{| N_k(\xi) |  }{  \sum_{o \in N_k(\xi)} \max(\textit{k-dist}(o),\textit{dist}(\xi,o))}
	\end{align*}
	is the local reachability density which measures the geometric dispersion of the $k$-nearest neighborhood.
	LOF effectively finds the instances fall in sparse regions of data.
%	In the experiments, we use $k=100$.
	}
	% INDEP-PRODUCT
	\item{\textit{Conditional outlier detection with $d$ independent response models} (I-PROD) -- 
		We apply \cite{Hauskrecht:2013} to the multivariate conditional setting by learning $d$ independent conditional probability models $P(Y_i|\mathbf{X})$ ($Y_i$ is not dependent on other output variables) and scoring based on the product of their estimates (Equation (\ref{eq:oscore_prod})).
		We refer to this method as I-PROD.
	}
	% MCODE-PRODUCT
	\item{\textit{MCODE without weighting} (M-PROD) (Equation (\ref{eq:oscore_prod})}
		%  -- Our proposed approach without using the reliability weights (Equation (\ref{eq:oscore_prod}))
	% MCODE-NONPARAM
	\item{\textit{MCODE with Reliability Weights} (M-RW) (Equation (\ref{eq:oscore}))}
	% MCODE-LOCAL_NONPARAM
	\item{\textit{MCODE with Local Reliability Weights} (M-LRW) (Equation (\ref{eq:oscore_local}))}
\end{itemize}
To obtain data models in I-PROD, M-PROD, M-RW, and M-LRW, we use $L_2$-penalized logistic regression and choose their regularization parameters by cross validation.
In LOF and M-LRW, we set the number of neighbors $k=100$.

% table:datasets
\begin{table}[t]
\centering
\resizebox{.485\textwidth}{!}{
	\begin{tabular}{ c  c  c  c  c }
		\hline\\[-.85em]
		\multirow{2}{*}{\textbf{Dataset}} &  \multirow{2}{*}{\textit{N \!/\! m \!/\! d}}  & \multirow{2}{*}{\textbf{Domain}}  & \multicolumn{2}{c}{\textbf{Value Description}} \\
		 &   &   & \textbf{Context} & \textbf{Response}\\
		%\textbf{Dataset} & \textit{N \!/\! m \!/\! d} & \textbf{Domain} & \textbf{Context} & \textbf{Response}\\
		\hline\\[-.85em]
		Mediamill & 43,907 \!/\! 120 \!/\! 101 & Video & Video frames & Concepts\\
		Enron & 1,702 \!/\! 1,001 \!/\! 53  & Text & Emails & Properties\\
		Bibtex & 7,395 \!/\! 1,836 \!/\! 159 & Text & Paper metadata & Topics\\
		Yahoo-business & 11,214 \!/\! 21,924 \!/\! 30 & Text & News articles & Topics\\
%		Yahoo-science & 6,428 \!/\! 37,187 \!/\! 40 & Text & News articles & Topics\\
		Yahoo-arts & 7,484 \!/\! 23,146 \!/\! 26 & Text & News articles & Topics\\
		Yeast & 2,417 \!/\! 103 \!/\! 14 & Biology & Genes & Functionalities\\								
		Genbase & 662 \!/\! 1,185 \!/\! 27 & Biology & Genes & Functionalities\\
		Birds & 645 \!/\! 276 \!/\! 19 & Sound & Bird songs & Species\\
%		Bookmarks & 87,856 & 2,150 & 208 & Text & Web document & Topics\\
%		Medical & 978 & 1,449 & 45 & Medical & Diagnosis notes & Diseases\\
		\hline\\[-.85em]
%		\multicolumn{6}{l}{* indicates bootstrapped datasets} &
	\end{tabular}}
	\caption{Dataset characteristics. ($N$: number of instances, $m$: input dimensionality, $d$: output dimensionality)}
	\label{table:datasets}
\end{table}

\subsection{Data}
\label{subsec:exp_data}

We use \textit{eight} public datasets with multi-dimensional input and output.\footnote{Datasets are available at {http://mulan.sourceforge.net} \cite{Tsoumakas:2010}.}
These are collected from various application domains, including 
sound recognition (\textit{Birds}), 
biology (\textit{Yeast}, \textit{Genbase}), 
text categorization (\textit{Yahoo} datasets, \textit{Bibtex}, \textit{Enron}), and
%medical diagnosis ({\textit{Medical}})
semantic video/image annotation (\textit{Mediamill}). 
%Each dataset consists of \textit{continuous} features (context) and associated \textit{binary} responses. 
Table \ref{table:datasets} summarizes the characteristics of the datasets, such as dataset size, data domain, and short descriptions of the input and output variables.

\begin{figure*}[t]
\begin{adjustwidth}{-.65in}{-.65in}
	\centering
	\subfigure[Mediamill (outlier dimensionality = $\{2.5, 5, 10\}\%$)]{\label{fig:r_mediamill}\includegraphics[width=1.05\textwidth]{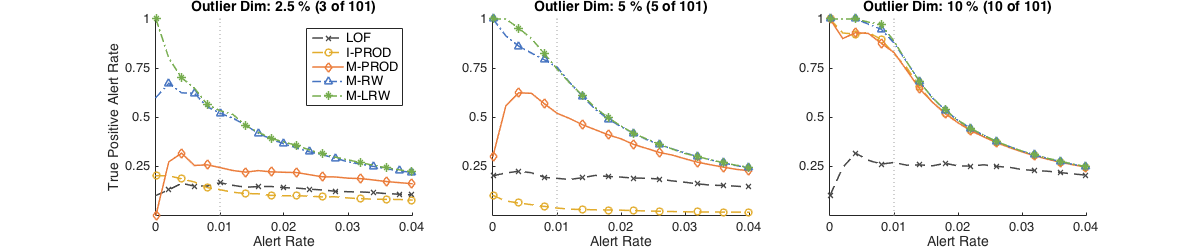}}\\
	\subfigure[Yahoo-arts (outlier dimensionality = $\{5, 10, 20\}\%$)]{\label{fig:r_yahoo_arts}\includegraphics[width=1.05\textwidth]{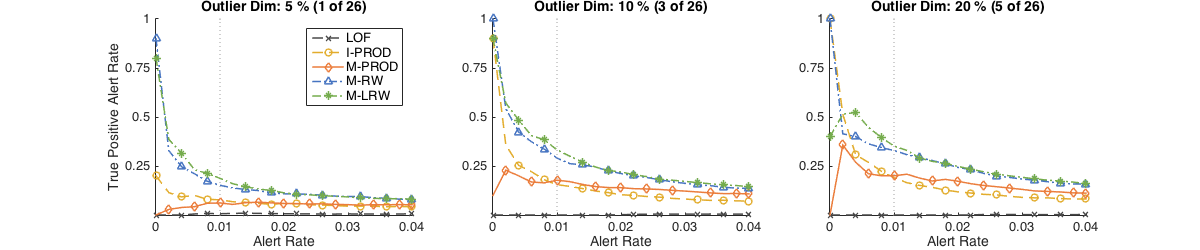}}
	\subfigure[Birds (outlier dimensionality = $\{5, 10, 20\}\%$)]{\label{fig:r_birds}\includegraphics[width=1.05\textwidth]{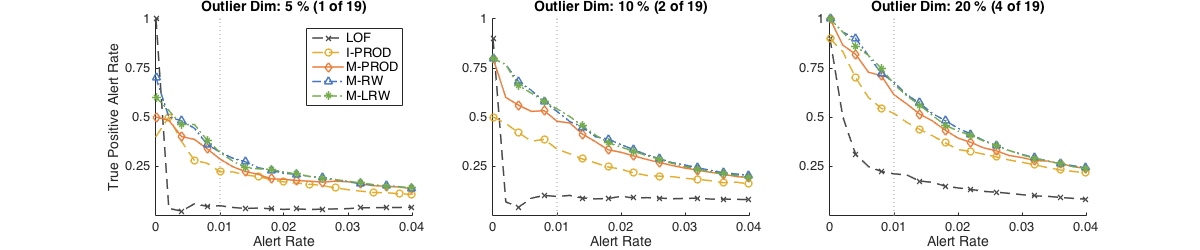}}
%	\subfigure[Yeast]{\label{fig:r}\includegraphics[width=1.15\textwidth]{fig/zz_yeast.png}}\\
%	\subfigure[Birds]{\label{fig:r}\includegraphics[width=1.15\textwidth]{fig/z_birds.png}}
%	\subfigure[Yahoo-business]{\label{fig:r}\includegraphics[width=1.15\textwidth]{fig/z_yahoo_business.png}}
	\caption{True positive alert rates at different alert rate ranging between 0 and 0.04.}
	\label{fig:tpar_results}
\end{adjustwidth}
\end{figure*}

\subsubsection{Simulating Outliers}
\label{subsubsec:exp_outlier}

For the purpose of our comparative evaluation, we simulate multivariate conditional outliers by perturbing the output space of data.
There are two parameters in our  simulation process: 
\textit{Outlier ratio} specifies how many outliers per simulation are injected. 
We set this parameter to $1\%$ throughout the experimental study. 
\textit{Outlier dimensionality} specifies how many output dimensions of an outlier to be perturbed.
We vary this parameter relative to the dimensionality of the output by perturbing $\{2.5, 5, 10, 20\}\%$ of outputs.
To summarize, we simulate outliers as:
\begin{enumerate}[leftmargin=*]
    \setlength\itemsep{-0.1em}
    \item{In each dataset, select $1\%$ of instances uniformly at random}
    \item{For each of the selected instances, perturb the values of $\{2.5, 5, 10, 20\}\%$ of the output dimensions (\ie, $y_\text{perturbed} = |y_\text{original}-1|$) uniformly at random}
\end{enumerate}
%Note that the injected outliers are analogous to unusual or incorrect input-output associations in each application domain (see Table \ref{table:datasets}).
%For example, in bird song identification (the \textit{Birds} dataset), an outlier instance represents inaccurate assignments of species identifiers to an acoustic waveform.
We would like to stress that all methods (including their model building and detection stages) are always run on data with injected outliers. That is, we never learn a model on the original (unperturbed) data and detect outliers on the simulated (perturbed) data. Such an design would be unrealistic since we do not know ahead of time what data instances to remove to learn a model. 

Note that the simulated outliers can be analogous to the errors or mistakes in each application domain.
%%%% For example, in bird sound identification, 
For example, in semantic video/image annotation, perturbed output values can be perceived as inaccurate subject labels.

\subsection{Evaluation Metrics} 

We use \textit{true positive alert rate} (TPAR) as our evaluation metric:
$$\textit{TPAR} = \textit{(True positive outliers)} / \textit{(Predicted outliers)}$$
TPAR (or precision) measures the percentage of instances with perturbation in the total number of instances detected by the methods.
%We evaluate TPAR at different alert rate (detection threshold).
% (Figure \ref{fig:tpar_results})
%averaged TPAR
We assess TPAR in two ways: 
We first evaluate TPAR at different alert rate (detection threshold) and analyze the quality of outlier scores 
(see Figure \ref{fig:tpar_results}).
We also measure the \textit{Averaged TPAR} (ATPAR) in $[0, 0.01]$ range, which coincides with the outlier ratio in our experiment setting.
For both TPAR and ATPAR, higher is better.

\begin{table*}[]
\centering
\resizebox{1\textwidth}{!}{
\renewcommand{\arraystretch}{1.09}
\begin{tabular}{c|ccclcccccclccc}
\hline
                       & \multicolumn{1}{l}{} & \multicolumn{6}{c}{Outlier dimensionality = 2.5\%}                                             & \multicolumn{1}{l}{} & \multicolumn{6}{c}{Outlier dimensionality = 5.0\%}                                             \\ \cline{3-8} \cline{10-15} 
\textit{ATPAR} & \multicolumn{1}{l}{} & \multicolumn{2}{c}{Baselines}     &  & \multicolumn{3}{c}{MCODE}                           & \multicolumn{1}{l}{} & \multicolumn{2}{c}{Baselines}     &  & \multicolumn{3}{c}{MCODE}                           \\ \cline{3-4} \cline{6-8} \cline{10-11} \cline{13-15} 
                       &                      & LOF             & I-PROD          &  & M-PROD          & M-RW            & M-LRW           &                      & LOF             & I-PROD          &  & M-PROD          & M-RW            & M-LRW           \\ \cline{1-8} \cline{10-15} 
Mediamill      &  & 0.14 $\pm$ 0.16 & 0.17 $\pm$ 0.09          &  & 0.26 $\pm$ 0.17          & \textbf{0.61 $\pm$ 0.12} & \textbf{0.69 $\pm$ 0.09} &  & 0.20 $\pm$ 0.17 & 0.06 $\pm$ 0.05          &  & 0.57 $\pm$ 0.14          & \textbf{0.85 $\pm$ 0.05} & \textbf{0.90 $\pm$ 0.04} \\
Enron          &  & 0.01 $\pm$ 0.03 & 0.12 $\pm$ 0.19          &  & 0.11 $\pm$ 0.11          & 0.06 $\pm$ 0.11          & 0.05 $\pm$ 0.09          &  & 0.01 $\pm$ 0.03 & 0.15 $\pm$ 0.22          &  & 0.20 $\pm$ 0.17          & 0.17 $\pm$ 0.22          & 0.21 $\pm$ 0.26          \\
Bibtex         &  & 0.00 $\pm$ 0.00 & 0.25 $\pm$ 0.28          &  & 0.32 $\pm$ 0.30          & 0.27 $\pm$ 0.29          & 0.33 $\pm$ 0.30          &  & 0.00 $\pm$ 0.01 & \textbf{0.44 $\pm$ 0.27} &  & \textbf{0.49 $\pm$ 0.28} & \textbf{0.47 $\pm$ 0.28} & \textbf{0.51 $\pm$ 0.27} \\
Yahoo-business &  & 0.01 $\pm$ 0.02 & 0.13 $\pm$ 0.06          &  & \textbf{0.21 $\pm$ 0.10} & \textbf{0.36 $\pm$ 0.09} & \textbf{0.38 $\pm$ 0.07} &  & 0.01 $\pm$ 0.03 & 0.25 $\pm$ 0.08          &  & \textbf{0.43 $\pm$ 0.11} & \textbf{0.56 $\pm$ 0.08} & \textbf{0.58 $\pm$ 0.07} \\
Yahoo-arts     &  & -               & -                        &  & -                        & -                        & -                        &  & 0.00 $\pm$ 0.01 & 0.11 $\pm$ 0.07          &  & 0.04 $\pm$ 0.04          & \textbf{0.26 $\pm$ 0.06} & \textbf{0.29 $\pm$ 0.08} \\
%Yeast          &  & -               & -                        &  & -                        & -                        & -                        &  & 0.08 $\pm$ 0.07 & 0.04 $\pm$ 0.06          &  & 0.45 $\pm$ 0.12          & \textbf{0.65 $\pm$ 0.06} & \textbf{0.64 $\pm$ 0.05} \\
Genbase        &  & -               & -                        &  & -                        & -                        & -                        &  & 0.05 $\pm$ 0.08 & \textbf{0.93 $\pm$ 0.05} &  & \textbf{0.93 $\pm$ 0.06} & \textbf{0.94 $\pm$ 0.06} & \textbf{0.95 $\pm$ 0.06} \\
Birds          &  & -               & -                        &  & -                        & -                        & -                        &  & 0.04 $\pm$ 0.08 & \textbf{0.34 $\pm$ 0.22} &  & \textbf{0.39 $\pm$ 0.25} & \textbf{0.45 $\pm$ 0.21} & \textbf{0.46 $\pm$ 0.22} 
\\ \hline
\multicolumn{1}{l}{}                        & \multicolumn{1}{l}{} & \multicolumn{1}{l}{} & \multicolumn{1}{l}{} &  & \multicolumn{1}{l}{} & \multicolumn{1}{l}{} & \multicolumn{1}{l}{} & \multicolumn{1}{l}{} & \multicolumn{1}{l}{} & \multicolumn{1}{l}{} &  & \multicolumn{1}{l}{} & \multicolumn{1}{l}{} & \multicolumn{1}{l}{} \\ \hline
                       & \multicolumn{1}{l}{} & \multicolumn{6}{c}{Outlier dimensionality = 10.0\%}                                             & \multicolumn{1}{l}{} & \multicolumn{6}{c}{Outlier dimensionality = 20.0\%}                                             \\ \cline{3-8} \cline{10-15} 
\textit{ATPAR} & \multicolumn{1}{l}{} & \multicolumn{2}{c}{Baselines}     &  & \multicolumn{3}{c}{MCODE}                           & \multicolumn{1}{l}{} & \multicolumn{2}{c}{Baselines}     &  & \multicolumn{3}{c}{MCODE}                           \\ \cline{3-4} \cline{6-8} \cline{10-11} \cline{13-15} 
                       &                      & LOF             & I-PROD          &  & M-PROD          & M-RW            & M-LRW           &                      & LOF             & I-PROD          &  & M-PROD          & M-RW            & M-LRW           \\ \cline{1-8} \cline{10-15} 
Mediamill      &  & 0.27 $\pm$ 0.16 & \textbf{0.92 $\pm$ 0.03} &  & \textbf{0.91 $\pm$ 0.04} & \textbf{0.97 $\pm$ 0.03} & \textbf{0.98 $\pm$ 0.03} &  & 0.30 $\pm$ 0.12 & \textbf{0.99 $\pm$ 0.02} &  & \textbf{0.99 $\pm$ 0.01} & \textbf{1.00 $\pm$ 0.01} & \textbf{1.00 $\pm$ 0.00} \\
Enron          &  & 0.03 $\pm$ 0.05 & \textbf{0.21 $\pm$ 0.24} &  & \textbf{0.23 $\pm$ 0.17} & \textbf{0.31 $\pm$ 0.27} & \textbf{0.39 $\pm$ 0.29} &  & 0.02 $\pm$ 0.04 & 0.26 $\pm$ 0.32          &  & 0.35 $\pm$ 0.21          & \textbf{0.62 $\pm$ 0.24} & \textbf{0.78 $\pm$ 0.17} \\
Bibtex         &  & 0.00 $\pm$ 0.01 & \textbf{0.70 $\pm$ 0.20} &  & \textbf{0.70 $\pm$ 0.18} & \textbf{0.71 $\pm$ 0.20} & \textbf{0.72 $\pm$ 0.17} &  & 0.00 $\pm$ 0.00 & \textbf{0.88 $\pm$ 0.11} &  & \textbf{0.86 $\pm$ 0.12} & \textbf{0.88 $\pm$ 0.11} & \textbf{0.87 $\pm$ 0.11} \\
Yahoo-business &  & 0.01 $\pm$ 0.01 & 0.32 $\pm$ 0.10          &  & \textbf{0.42 $\pm$ 0.13} & \textbf{0.57 $\pm$ 0.06} & \textbf{0.57 $\pm$ 0.07} &  & 0.01 $\pm$ 0.02 & \textbf{0.36 $\pm$ 0.13} &  & 0.25 $\pm$ 0.09          & \textbf{0.39 $\pm$ 0.05} & \textbf{0.41 $\pm$ 0.04} \\
Yahoo-arts     &  & 0.00 $\pm$ 0.00 & 0.29 $\pm$ 0.05          &  & 0.18 $\pm$ 0.07          & \textbf{0.44 $\pm$ 0.06} & \textbf{0.47 $\pm$ 0.05} &  & 0.00 $\pm$ 0.00 & 0.36 $\pm$ 0.05          &  & 0.26 $\pm$ 0.07          & 0.39 $\pm$ 0.06          & \textbf{0.45 $\pm$ 0.07} \\
Yeast          &  & 0.08 $\pm$ 0.07 & 0.04 $\pm$ 0.06          &  & 0.45 $\pm$ 0.11          & \textbf{0.64 $\pm$ 0.06} & \textbf{0.63 $\pm$ 0.05} &  & 0.13 $\pm$ 0.09 & 0.17 $\pm$ 0.11          &  & \textbf{0.52 $\pm$ 0.08} & \textbf{0.56 $\pm$ 0.07} & \textbf{0.54 $\pm$ 0.08} \\
Genbase        &  & 0.06 $\pm$ 0.11 & \textbf{0.96 $\pm$ 0.06} &  & \textbf{0.96 $\pm$ 0.04} & \textbf{0.98 $\pm$ 0.02} & \textbf{0.98 $\pm$ 0.02} &  & 0.03 $\pm$ 0.09 & \textbf{0.98 $\pm$ 0.03} &  & \textbf{0.96 $\pm$ 0.03} & \textbf{0.98 $\pm$ 0.02} & \textbf{0.98 $\pm$ 0.03} \\
Birds          &  & 0.07 $\pm$ 0.11 & \textbf{0.42 $\pm$ 0.31} &  & \textbf{0.56 $\pm$ 0.24} & \textbf{0.66 $\pm$ 0.18} & \textbf{0.66 $\pm$ 0.19} &  & 0.32 $\pm$ 0.22 & \textbf{0.67 $\pm$ 0.25} &  & \textbf{0.78 $\pm$ 0.19} & \textbf{0.85 $\pm$ 0.12} & \textbf{0.84 $\pm$ 0.13} 
 \\ \hline
\end{tabular}
}
\caption{Averaged true positive alert rate in [0, 0.01]. Numbers shown in bold indicate the best results on each experiment set (by paired t-test at $\alpha\!=\!0.05$).}
\label{table:aucprec}
\end{table*}

\subsection{Results}
\label{subsec:results}

Figure \ref{fig:tpar_results} and Table \ref{table:aucprec} show the performance of the five compared methods.
All results are obtained from \textit{ten} repeats.

Figures \ref{fig:r_mediamill}, \ref{fig:r_yahoo_arts}, and \ref{fig:r_birds} present the results on three datasets (\textit{Mediamill}, \textit{Yahoo-arts}, and \textit{Birds}) for different outlier dimensions.
Each figure illustrates the TPARs of all methods; 
X-axes show the alert rate, ranging between 0 and 0.04; Y-axes show TPAR.
The vertical gray line at alert rate $= 0.01$ indicates where the alert rate is equal to the injected outlier ratio.

In general, TPARs improve as the outlier dimensionality increases, because outliers with larger perturbations are easier to detect.
Comparing the conditional outlier detection approaches (I-PROD, M-PROD, M-RW, and M-LRW) with the unconditional approach (LOF), the conditional approaches are clear winners as the conditional methods outperform LOF in most cases.
This shows the advantages of the conditional outlier detection approaches in addressing the problem. 
Only exceptions are I-PROD on \textit{Mediamill} when outlier dimensionality is low.
This is because I-PROD does not consider the dependence relations among the output variables.
Such advantages in modeling the inter-dependences of the outputs are consistently observed as M-PROD outperforms I-PROD in most experiments.

To show the benefits of our reliability weights, we analyze the performance of M-RW and M-LRW in comparison to that of M-PROD.
An interesting point is that M-RW and M-LRW not only improve the performance drastically, but also make TPARs stable.
This confirms that our reliability weighting methods can effectively estimate the quality of the models, and the resulting weights are useful in outlier scoring.
Lastly, although M-LRW does not show much improvement from M-RW compared to the other key components of MCODE that we have discussed, the local weights seem to make M-RW even more stable as shown with \textit{Mediamill} and \textit{Yahoo-arts}.

Table \ref{table:aucprec} summarizes the results on all eight datasets in terms of ATPAR at 0.01.
The table consists of four sections grouped by different values of outlier dimensionality ($\{2.5, 5, 10, 20\}\%$).
We do not report the results on the first four datasets (\textit{Birds}, \textit{Yeast}, \textit{Genbase}, and \textit{Yahoo-arts}) for outlier dimensionality $= 2.5\%$ (for \textit{Yeast}, 2.5\% and 5.0\%) because the output dimensionality ($d$) is too small.
The best performing methods on each experiment are shown in bold.

The results confirms the conclusions that we have drawn with Figure \ref{fig:tpar_results}.
One interesting point is that LOF shows exceptionally high (compared with its performance on other datasets) ATPAR on \textit{Mediamill}.
This is because the dataset has a similar number of input and output variables; hence, as outlier dimensionality increases, the simulated outliers become like unconditional outliers.

% CONCLUSION
\section{Conclusions}
\label{sec:concl}

In this work, we introduced and tackled multivariate conditional outlier detection, a special type of the conditional outlier detection problem.
We briefly reviewed existing research and motivated this new type of outlier detection problem.
We presented our novel outlier detection framework that analyzes and detects abnormal input-output associations in data using a decomposable conditional probabilistic model that is learned from all data instances.
We discussed how to obtain an efficient data representation and accurate conditional probability estimates of observed input-output pairs, using the probabilistic structured data modeling approach.
Motivated by the Brier score, we developed present two outlier scoring methods by analyzing the reliability of probability estimates.
%Since components of this model can vary in their quality, we combine them with the help of weights reflecting their reliability in assessment of outliers.
%We study two ways of calculating the component weights: global that relies on all data, and local that relies only on instances similar to the target instance.
Through the experimental results, we demonstrated the ability of our framework to successfully identify multivariate conditional outliers.

\bibliographystyle{plain}

\end{document}